\documentclass[10pt,twocolumn,letterpaper]{article}

\usepackage[pagenumbers]{cvpr} 

\usepackage{graphicx}
\usepackage{amsmath}
\usepackage{amssymb}
\usepackage{booktabs}

\usepackage{comment}
\excludecomment{tcolorbox}
\excludecomment{nonsubmit}

\usepackage[pagebackref,breaklinks,colorlinks]{hyperref}

\usepackage[capitalize]{cleveref}
\crefname{section}{Sec.}{Secs.}
\Crefname{section}{Section}{Sections}
\Crefname{table}{Table}{Tables}
\crefname{table}{Tab.}{Tabs.}

\def\cvprPaperID{*****} 
\def\confName{CVPR}
\def\confYear{2022}

\begin{document}

\title{Template NeRF: Towards Modeling Dense Shape Correspondences \\ from Category-Specific Object Images}

\author{
Jianfei Guo$^{*1}$ \qquad
Zhiyuan Yang$^{*2}$ \qquad
Xi Lin$^2$ \qquad
Qingfu Zhang$^2$
\\
$\,^1$Xi'an Jiaotong University \qquad
$\,^2$City University of Hong Kong
\\
{\tt\small ffventus@gmail.com} \qquad
{\tt\small \{zhiyuan.yang, xi.lin, qingfu.zhang\}@cityu.edu.hk}
}

\twocolumn[{
\vspace{-3em}
\maketitle
\vspace{-2em}
\begin{center}
    \captionsetup{type=figure}
    \includegraphics[width = \textwidth]{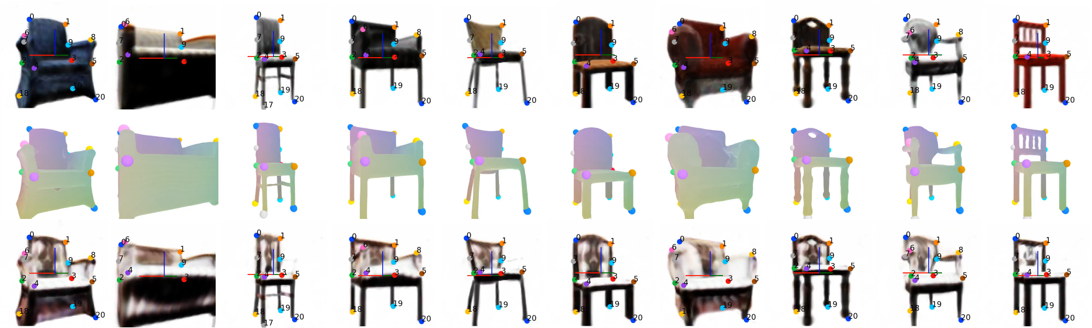}
    \captionof{figure}{
    \textbf{Results on ShapeNet chairs}~\cite{chang2015shapenet}.
    We show rendered images in a specific position after learning unsupervised 3D representation from a collection of multi-view posed images (first-row).
    Our model simultaneously and automatically reason dense shape correspondence across all the shapes.
    We extract the shapes by using Marching Cubes~\cite{lorensen1987marching} and depict the same correspondence by consistent colors (second-row).
    We demonstrate that the model is capable of transferring texture from a given instance's texture (third-row).
    To better compare with traditional discrete methods, we also manually annotate several keypoints on one object instance and compute their new locations on other instances, colored points as shown in all three rows. 
    }
    \label{fig:teaser}
\end{center}
}]

\renewcommand{\thefootnote}{}
\footnote{$*$ These authors contributed equally to this work.}

\begin{abstract}
We present neural radiance fields (NeRF) with templates, dubbed \textbf{Template-NeRF}, for modeling appearance and geometry and generating dense shape correspondences simultaneously among objects of the same category from only multi-view posed images, without the need of either 3D supervision or ground-truth correspondence knowledge. The learned dense correspondences can be readily used for various image-based tasks such as keypoint detection, part segmentation, and texture transfer that previously require specific model designs. Our method can also accommodate annotation transfer in a one or few-shot manner, given only one or a few instances of the category. Using periodic activation and feature-wise linear modulation (FiLM) conditioning, we introduce deep implicit templates on 3D data into the 3D-aware image synthesis pipeline NeRF. By representing object instances within the same category as shape and appearance variation of a shared NeRF template, our proposed method can achieve dense shape correspondences reasoning on images for a wide range of object classes. We demonstrate the results and applications on both synthetic and real-world data with competitive results compared with other methods based on 3D information.
\end{abstract}

\section{Introduction}

\begin{figure*}[ht]
\centering
	\includegraphics[width = \textwidth]{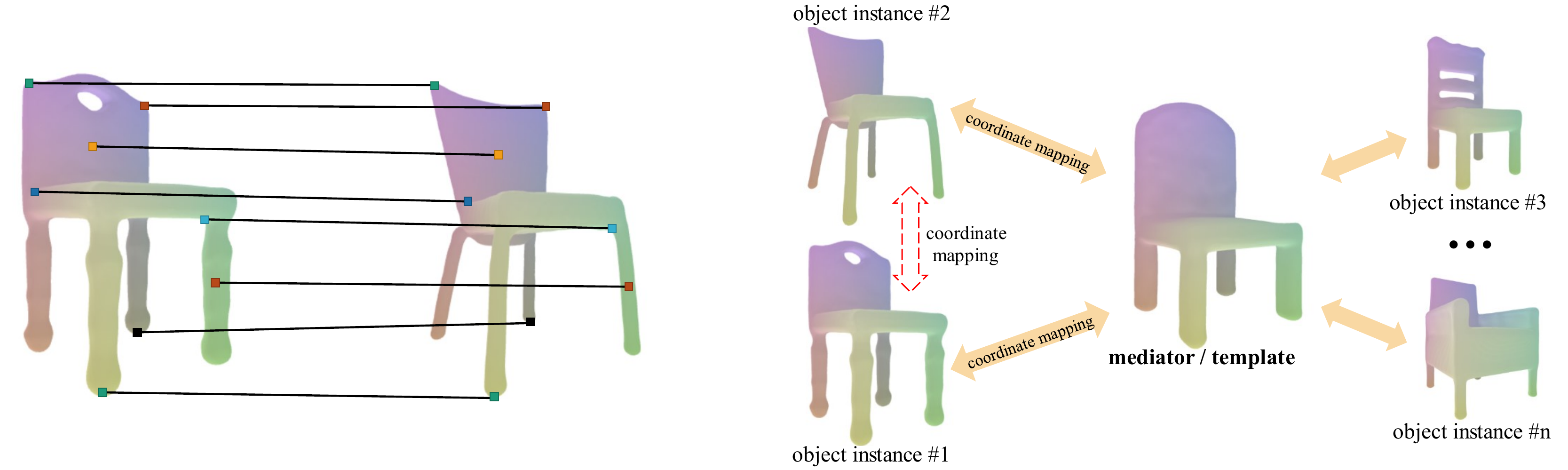}
    \caption{\textbf{Dense shape correspondence}, i.e., the continuous coordinate mapping relationship between surface points of two objects everywhere. Several pairs of points are manually selected to show their correspondences (left). If  one the coordinate mapping relationship between a shared template and each object has been estalished, the coordinate mapping among object instances can be obtained using the shared template as the mediator (right).}
\label{fig:Corr}
\end{figure*}


A fundamental assumption in computer vision community is that objects within the same category share some common shape features with semantic correspondences~\cite{blanz19993dmm, loper2015smpl, kulkarni2019canonical, zheng2020DIT, deng2020DIF}.
This relation should be dense by nature and can be seen as the generalization and densification of keypoints, serving as a stepping stone towards a better understanding of objects and inference of information.
It can be further modeled as the continuous coordinate mapping relation between surface points of two objects, which strictly ensures semantic alignment everywhere (e.g., keypoint and part surface alignment) and is often termed \textit{dense shape correspondences}, as shown in Figure~\ref{fig:Corr}.
Previous efforts~\cite{zhou2016learning, kulkarni2019canonical} in 2D domain focus on dense correspondence utilizing a cycle mechanism with an intermediate abstract 3D model.
Dense shape correspondences are more common in the settings where 3D ground-truth knowledge is available~\cite{halimi2019unsupervised, li2019usip, zeng2021corrnet3d}.

Major motivations behind our work are: (1) If we can directly infer dense shape correspondence from images in advance, annotations or modifications could be transferred across the instances.
The category-specific tasks (e.g., keypoint detection~\cite{zadeh2017convolutional, dong2019teacher, thewlis2019unsupervised, jakab2020self}, texture transfer~\cite{mir2020learning}, and part segmentation~\cite{hung2019scops, larsson2019cross}) can be conducted in  one or few-shot manner using a generic model. As a result, the computational overhead can be
significantly reduced and interpretability can be acheived. (2) If the coordinate mapping relationship between a shared space (typically, a template shape) and each object instance's space has been established, the coordinate mapping among object instances can be obtained using the shared template as a mediator.

DIT~\cite{deng2020DIF} and DIF~\cite{zheng2020DIT} decompose implicit representations~\cite{park2019deepsdf} of 3D shapes into a template field and a warping/deformation field by learning from 3D datasets. The warping field maps points of object instance space into template space, thus coordinate mapping relationship of object instances can be obtained via nearest neighbor search in the template space. 
However, these works are based on 3D shapes, and there are still large gaps in applying their results to image-based tasks. 
Moreover, large category-specific 3D shape datasets are crucial but labor-intensive for these attempts.
Hence, we believe that it is important and necessary to reason dense shape correspondences directly from images. 

 
Recently, in the field of neural rendering and 3D-aware image synthesis, much effort has been made to learn implicit representations encoding both appearance and geometry from images without any 3D supervision~\cite{mildenhall2020nerf, kaizhang2020nerfplusplus, Lombardi:2019NV, tretschk2020nonrigid,  sitzmann2019srns, DVR}. 
These works have shown superior reconstruction results compared to traditional methods such as visual Structure-from-Motion~\cite{andrew2001multiple-sfm1, triggs1999bundle-sfm2, snavely2006photo-sfm3}.
Among these emerging technologies, Neural Radiance Fields (NeRF)~\cite{mildenhall2020nerf} has demonstrated its powerful representation abilities using simple multi-layer perception (MLP) networks.

NeRF maps spatial location and view direction into volume density (representing shapes) and emitted radiance values (representing appearance), and uses volume rendering to synthesis RGB images with extraordinary realism and details.
However, vanilla NeRF can only render novel views of a single object or a single scene. 
Following works~\cite{schwarz2020graf,chan2020pigan} condition NeRF-like network on the latent shape and appearance codes to form category-specific implicit representations, which learns shape and appearance of multiple objects of the same class from images leveraging a GAN-based~\cite{Goodfellow2014GenerativeAN} structure.


In this paper, we combine category-specific NeRF with deep implicit templates.
We use a NeRF template to model common geometry, structure, and appearance of the objects within the same category.
Following a deformation field, a volume density correction field and appearance code conditioning account for geometry, structure and appearance variance of a specific object, respectively.
The learned explicit shape correspondences can be used  for many applications. We demonstrate its applications in the one-shot semantic labeling task. Users only need to annotate one instance to obtain annotations across the whole category. The annotations take any form of key-points, key-curves, key-surfaces, or even key-areas. 

Our contributions in this paper are summarized as follows:
\begin{itemize}
\item We introduce deep implicit templates to NeRF with careful design, allowing for explicit dense shape correspondences reasoning using only posed images.
\item We propose a modified version of the FiLM-SIREN layer to achieve high-quality results.
\item We demonstrate novel applications of dense correspondences using only image supervision, such as one- / few-shot keypoint detection and texture transfer.
\end{itemize}

\section{Related Work}

{\bf Implicit Neural Representations and Rendering.}
Implicit Neural Representations~\cite{mescheder2019occupancy, park2019deepsdf, chen2019learning, gropp2020implicit, xu2019disn, tretschk2020patchnets} have been exploded in recent years by using implicit functions to represent 3D geometry objects or scenes.
Compared with the conventional approaches based on voxel grids~\cite{wu20153d,wu2018learning} or meshes~\cite{ranjan2018generating, wang2018pixel2mesh} which discretize space and are restricted in topology, implicit representations provide a compact and continuous mechanism.
This advantage also makes it possible to be trained by only 2D images using neural rendering. 
Neural rendering projects a 3D neural representation into multiple 2D images which could backpropagate the reconstruction error to optimize.
DeepSDF~\cite{park2019deepsdf} proposes to learn an implicit function where the output of the network represents the signed distance of the point to its nearest surface.
NeRF~\cite{sitzmann2020siren} approximates a continuous 5D scene representation with a multilayer perceptron (MLP) network $F_\Theta$, which outputs volume density $\sigma$ and emitted color $\mathbf{c}$ given a 3D location $\mathbf{x} = (x, y, z)$ and a 2D viewing direction $\mathbf{d} = (\theta, \phi)$.

NeRF has been broadly extended with many enhancements, such as dealing with more extreme light conditions~\cite{martinbrualla2020nerfinthewild}, achieving more realism in reflectance~\cite{nerv2020, bi2020neuralreflectancefield}, and dealing with dynamic or deformable scenes~\cite{park2020nerfies, pumarola2020d-nerf, li2020nsff, tretschk2020nonrigid}. 
Several works~\cite{chan2020pigan, schwarz2020graf} have developed NeRF as a category-specific representation using generative adversarial networks, generalizing well across instances, different from vanilla NeRF which only captures a single scene. 

{\bf Shape Correspondences.}
Recent works mainly focus on learning shape correspondences without manual supervision while the previous techniques~\cite{zhang2014facial, liu2010sift, choy2016universal, guler2018densepose} need poses or face keypoints annotations.
From the perspective of 2D features, 
\cite{rocco2018end, rocco2017convolutional} learn a parametric warping function for one to another related images,
and \cite{thewlis2017unsupervised, wiles2018self, thewlis2017unsupervised, thewlis2019unsupervised, jakab2020self} learn equivariant embeddings for matching the correspondences.
However, these methods are restricted to highly homogeneous training data in the structure, which largely simplifies the difficulty of the tasks.
Compared with directly reasoning correspondence on 2D features,
learning on 3D structures as an intermediate medium is more powerful~\cite{you2020semantic, kulkarni2019canonical, zhou2016learning}.
\cite{you2020semantic} leverages a 2D-3D-2D cycle to ease the self-occlusion and enhance spatial relationships, but it needs to transfer keypoints using existing 3D datasets.
\cite{zhou2016learning, kulkarni2019canonical} utilize a cycle consistency mechanism as a supervisory signal to train and generalize the keypoint detection to segmentation mask.
\cite{kulkarni2020articulation} further infers the articulation and pose.
However, They need mask labels to separate the background and foreground pixels for reasoning correspondences.

Learning from 3D dataset, element-based or mesh-based methods~\cite{groueix20183d, groueix2018atlasnet, genova2019learning, deprelle2019learning} can easily establish dense correspondences.
After the seminal dense shape model work of 3D morphable model (3DMM)~\cite{blanz19993dmm} is proposed for the human face, it has inspired various other domains~\cite{loper2015smpl, khamis2015learning}.

Reasoning dense shape correspondences of complex objects is still a challenge, especially in the settings of learning from only 2D supervision without 3D knowledge.
Recently, \cite{zheng2020DIT, deng2020DIF} propose to decompose the implicit representations into a template field and a warping/deformed field. 
Inspired by them but transferred to 2D scope, our method directly learns a dense correspondence from images which is able to guarantee global smoothness and arbitrary resolution in the morphing result.

{\bf 3D-Aware Image Synthesis.} In the field of 2D machine vision, deep generative models have achieved considerable success with the introduction of Generative Adversarial Networks (GANs)~\cite{Goodfellow2014GenerativeAN}.
3D-aware synthesis is also needed because our real-world objects are three-dimensional and 3D information also brings richer perception.
Though several approaches~\cite{tran2017disentangled, tian2018cr} have been tried to disentangle pose and identity, these 2D generations still struggle to synthesize novel view images with identity consistent at high quality. 
To address this problem, 3D representation models with a generative mechanism have exploded in recent years.

HoloGAN~\cite{nguyen2019hologan} and BlockGAN~\cite{nguyen2020blockgan} use voxelized feature-grid representations with a learnable projection form 3D to 2D.
GRAF~\cite{schwarz2020graf} is a conditional variant of NeRF which makes image synthesis controllable.
GIRAFFE~\cite{niemeyer2020giraffe} first brings compositional structure into the generative model to handle multi-object scenes.
The latest work $\pi$-GAN~\cite{chan2020pigan} combines SIREN-layer, styleGAN-inspired mapping network and progressive training strategy to achieve high-quality results.

\section{Method}

\begin{figure*}[tb]
	\centering
	\includegraphics[width = \linewidth]{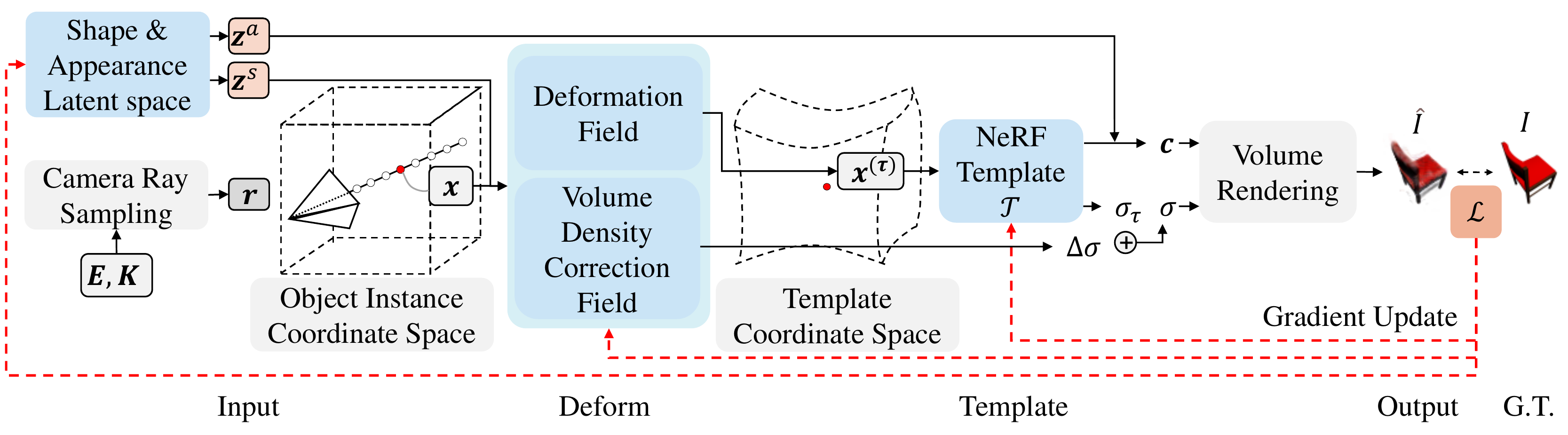}
	\caption{\textbf{Overall framework of Template-NeRF.} Our model consists of three major trainable components: 1) Shape / Appearance conditional code, 2) Deformation and Volume Density Correction field, and 3) Template NeRF model. Given a conditional shape code, a number of 3D points $\mathbf{x}$ sampled along the camera ray will be warped into a shared template nerf model. We account for the geometry variance by the deformation field and the structure variance by the volume density correction field. The conditional appearance code will affect the emitted radiance field. We render the synthesis image by the output of the template NeRF model together with the variance.
	}
	\label{fig:Framework}
\end{figure*}

\begin{figure}[b]
	\centering
		\centering
		\includegraphics[width = 1\linewidth]{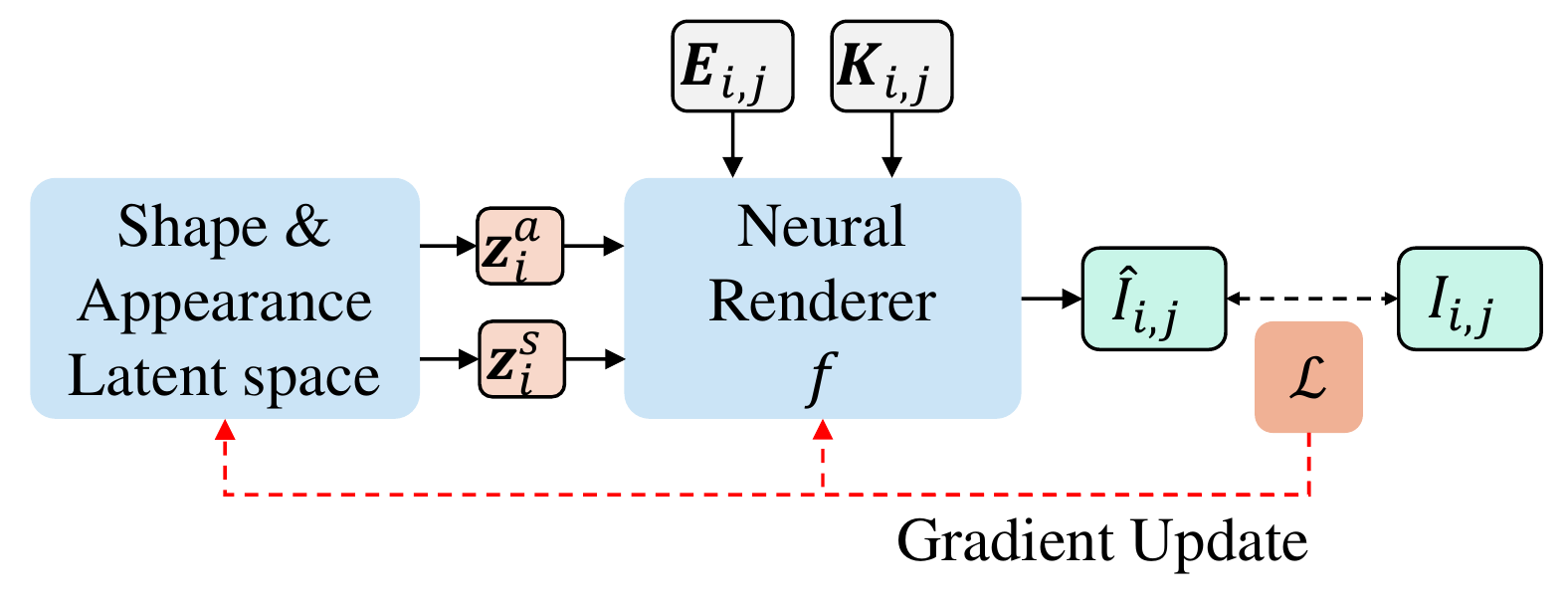}
		\caption{\textbf{Framework of the simple Auto-Decoder NeRF without template.} Different from Auto-Encode, the learning process directly starts from randomly sampled latent codes. Then we update the parameters of the latent codes and the neural networks simultaneously by minimizing the photometric reconstruction.}
		\label{fig:AutoDecoderNeRF}
\end{figure}

\subsection{Formulation and Overview}

Consider a training set $\{\mathcal{O}_i\}_{i=1}^N$ of $N$ object instances within a single category. Each instance $\mathcal{O}_i$ is comprised of tuples $\mathcal{O}_i=\{\mathcal{I}_j, E_j, K_j\}_{j=1}^{M_i}$ from $M_i$ sparse observations of an object $\mathcal{O}_i$, where  $\mathcal{I}_j \in \mathbb{R}^{H \times W \times 3}$ is a posed image with its respective extrinsic $E_j = [R|t] \in \mathbb{R}^{3 \times 4}$ and intrinsic $K_j \in \mathbb{R}^{3 \times 3}$ camera matrices. 
Our goal is to model the appearance, geometry and dense shape correspondences simultaneously across object instances $\{\mathcal{O}_i\}_{i=1}^{N}$.
Note that our approach does \textit{NOT} require either information about 3D object geometry or ground truth correspondence annotations. It only needs the posed 2D RGB images $\mathcal{O}_i=\{\mathcal{I}_j, E_j, K_j\}_{j=1}^{M_i}$ which is come-at-able and makes it more practical for real-world applications.

We set up an auto-decoder pipeline to jointly learn the latent shape codes $\{\mathbf{z}_i^s \in \mathbb{R}^{s}\}_{i=1}^{N}$, the appearance codes $\{\mathbf{z}_i^a \in \mathbb{R}^{a}\}_{i=1}^{N}$ and the weights of neural renderer $f$. The entire inference rendering process can be written as follows:
\begin{equation}
	\hat{\mathcal{I}}_{i,j}=f\left(\mathbf{z}^s_i, \mathbf{z}^a_i, E_{i,j}, K_{i,j}\right),
\end{equation}
where $f$ render an estimated image observation $\hat{\mathcal{I}}_{i,j}$ of object $\mathcal{O}_i$, conditioned on $\mathbf{z}_i^s$, $\mathbf{z}_i^a$ and camera parameters $E_{i,j}, K_{i,j}$ of the current view $j$.
To learn the optimal model parameters and latent codes, we minimize the reconstruction loss between the rendered images $\hat{\mathcal{I}}_{i,j}$ and the ground truth images $\mathcal{I}_{i,j}$, along with proper regularization. 

As shown in Figure~\ref{fig:AutoDecoderNeRF}, the above formation incorporates the category-specific representation into the vanilla NeRF structure, which we term \textit{Auto-Decoder NeRF}. Auto-Decoder NeRF is capable of learning shapes and appearances of different objects, but not dense shape correspondence.
This is similar to \cite{schwarz2020graf, chan2020pigan, niemeyer2020giraffe}, where the shape / appearance variance of individual objects is implicitly reflected by the changes in volume density $\sigma$ / emitted radiance values $\mathbf{c}$, respectively.

Following the ideas of analysis by synthesis, to reason about dense shape correspondences, we further decompose the neural model $f$ into a NeRF template $\mathcal{T}$, a deformation field, and a volume density correction field. 
Inspired by \cite{deng2020DIF, zheng2020DIT}, the shape/appearance of individual objects can be alternatively viewed as shape/appearance variation from a NeRF template model that captures the common shape and appearance features of objects within the same category.
The deformation field and volume density correction field, both conditioned on the latent shape codes, cover the geometry and structure aspect of shape variance, respectively.

Concerning the geometry changes, the deformation field warps the object instance's space into the template space, which can be viewed as deforming the object shape into the template shape.
This warping function establishes the coordinate mapping relationship between object instances and the template, through which dense shape correspondences of different objects are obtained via nearest neighbor search among their warped surface points in the template space.

The overall framework is shown in Figure~\ref{fig:Framework}. In the following sections, we will illustrate the details of the NeRF template model, the appearance and the shape variance model (deformation field and volume density correction field), the training losses and the learning procedure.
For details of NeRF and its neural rendering process, we suggest readers refer to the original NeRF paper~\cite{mildenhall2020nerf}.

\subsection{NeRF Template and Appearance Variance}

The NeRF template $\mathcal{T}$ is shared across the same category of objective, and is jointly trained with the shape variance model and appearance variance model. It is parameterized as a multi-layer perceptron (MLP) that takes a 3D location $\mathbf{x} = (x, y, z)$ and 2D viewing direction $\mathbf{d} = (\theta, \phi)$ as input, and outputs volume density $\sigma$ and emitted radiance value $\mathbf{c}$ of the related spatial location.
By sampling $P$ points $\{ \mathbf{p_{i,j,k}} \}_{k=1}^{P}$ along the sampled camera ray $\mathbf{r}$ of the current view $E_{i,j}, K_{i,j}$ of the current object $\mathcal{O}_i$, and evaluating $\sigma$ and $\mathbf{c}$ on the sampled points, one can obtain the predicted RGB value $C(\mathbf{r})$ of the related sampled pixel through volume integration:
\begin{equation}
\begin{aligned}
	C(\mathbf{r}) = \int_{t_n}^{t_f} T(t) \cdot \sigma(r(t)) \cdot \mathbf{c}(r(t),d) \, {\rm d}t, \\ \text{where} \; T(t) = e^{-\int_{t_n}^t \sigma(r(s)){\rm d}s}.
\end{aligned}
\end{equation}
In the following content, we mainly discuss the computations on each point $\mathbf{p}_{i,j,k}$. To simplify the notations, we ignore $i$, $j$ and $k$ subscripts unless otherwise noted.

\begin{figure}[t]
\centering
\includegraphics[width=1\linewidth]{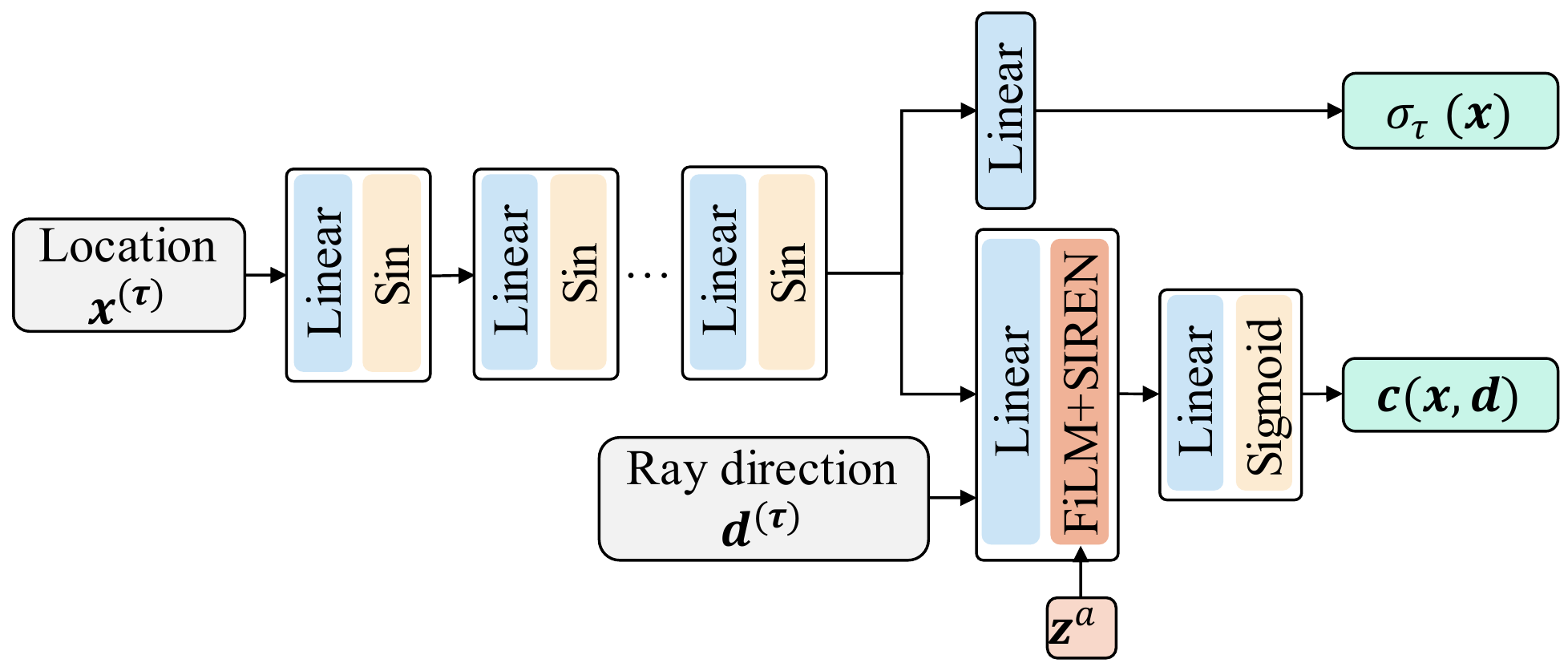}
\caption{The NeRF template and appearance variance model with sinusoidal activation and FiLM-SIREN layer.}
\label{fig:NeRF-template-appear}
\end{figure}

As suggested in SIREN~\cite{sitzmann2020siren}, using sinusoidal activation instead of ReLU on spatial implicit representation like NeRF~\cite{mildenhall2020nerf} or DeepSDF~\cite{park2019deepsdf} could lead to smoother representation results. Hence, we apply sinusoidal activation to each layer of the NeRF template except the last layer of volume density field and the last layer of emitted radiance field, as shown in Figure~\ref{fig:NeRF-template-appear}. 

To account for the shape variance of each object $\mathcal{O}_i$, the input coordinates are warped before passing to the NeRF template to cover the geometry variance, and a volume density correction is added to the original output of volume density of $\mathcal{T}$ to cover the structure variance.

To account for the appearance variance of each individual objects, we condition the emitted radiance branch of the NeRF template $\mathcal{T}$ using Feature-wise Linear Modulation (FiLM)~\cite{perez2018film} of SIREN~\cite{sitzmann2020siren} layer, which is termed the \textit{FiLM-SIREN} layer:
\begin{equation}
	\phi\left(\mathbf{h}_{i}\right)=\sin \left(\boldsymbol{\gamma}_{i} \times \mathbf{h}_{i}+\boldsymbol{\beta}_{i}\right),
\end{equation}
where $\mathbf{h}_{i}$ denotes the layer's input, $\times$ denotes the element-wise multiplication of two equally-sized vectors, ${\mathbf{\gamma}}_{i}$ denotes the frequencies and $\boldsymbol{\beta}_{i}$ denotes the phase shifts, both of which are conditioned on the latent $\mathbf{z}$ via a mapping network, similar with Style-GAN\cite{karras2019style}.

\begin{figure}[t]
\centering
\includegraphics[width=1\linewidth]{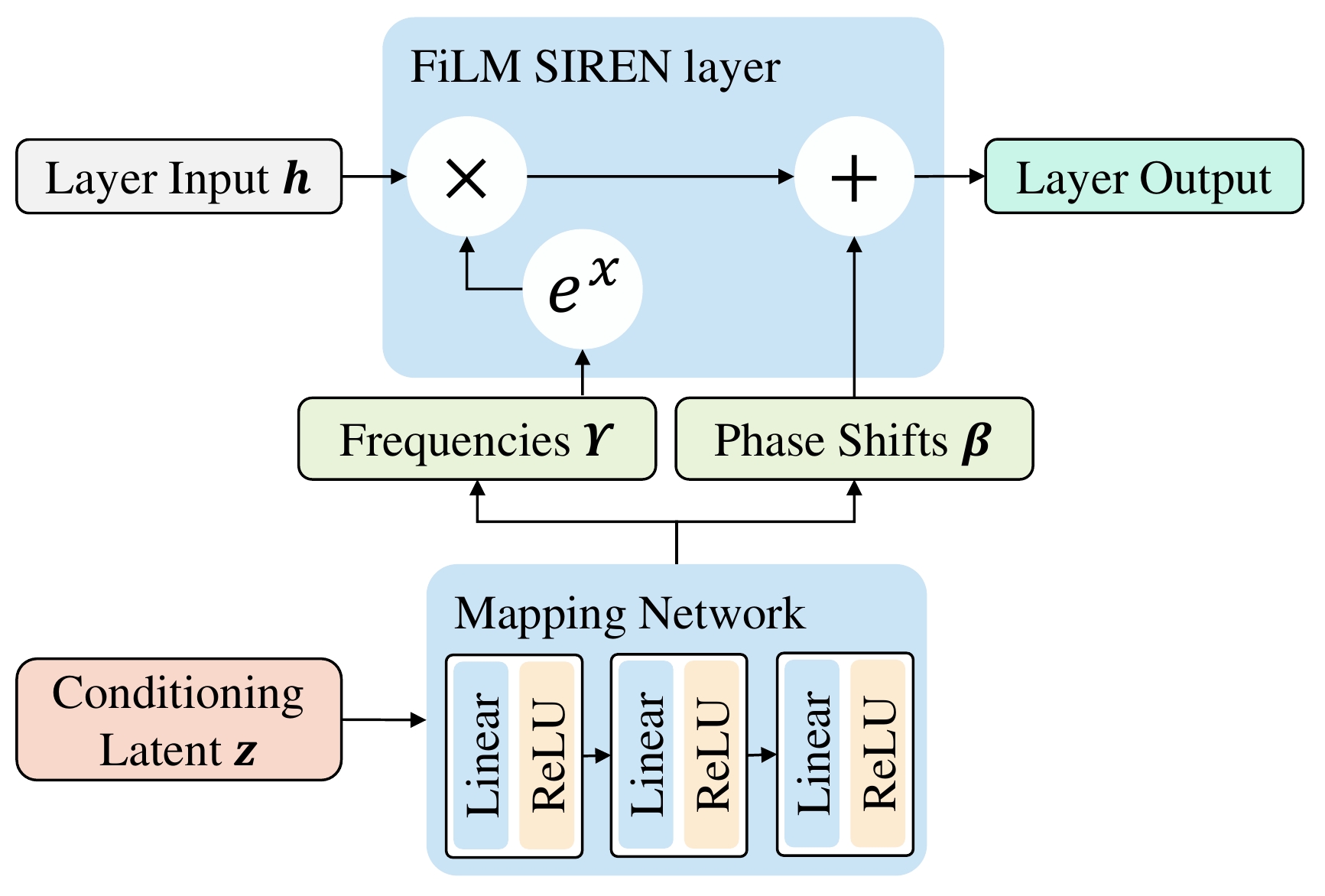}
\caption{The modified version of the FiLM-SIREN layer where we pass the frequencies $\gamma$ through an exponential function.}
\label{fig:film-siren}
\end{figure}

In practice, we find that directly multiplying frequencies vectors $\mathbf{\gamma}$ with the layer input $\mathbf{h}$ would lead to gradient vanishing, thus we pass the frequencies $\mathbf{\gamma}$ through element-wise exponential function $f(x)=e^x$ before the multiplication with the layer input, as shown in Figure~\ref{fig:film-siren}:
\begin{equation}
	\phi\left(\mathbf{h}_{i}\right)=\sin \left(e^{\boldsymbol{\gamma}_{i}} \times \mathbf{h}_{i}+\boldsymbol{\beta}_{i}\right),
	\label{equ:film-siren}
\end{equation}
which has achieved significant training speed boost up and convergence assurance.

\subsection{Shape Variance Model} 

Each shape of the object $\mathcal{O}_i$ can be viewed as a conditional variation from the shape of the NeRF template $\mathcal{T}$. Shape variation typically comprises geometry change (i.e., bending, scaling, and other topology-preserving deformations) and structural change (i.e., change of topology). We use a deformation field to account for geometry change and a volume density correction field for structural change.

We model the deformation as a spatial warping function $\mathcal{W}$ that maps spatial coordinates of individual objects space $\mathbf{x}^{(i)}\in\mathcal{O}_i$ to the locations in template space $\mathbf{x}^{(\tau)}\in\mathcal{O}_{\tau}$. The deformation field network computes the coordinate offsets, noted as $\mathcal{D}^{w}(\mathbf{x})$, conditioned on the shape latent code $\mathbf{z}_i^s$:
\begin{equation}
    \mathbf{x}^{(\tau)} = \mathcal{W}(\mathbf{x}^{(i)}, \mathbf{z}_i^s) = \mathcal{D}^{w}|_{\mathbf{z}_i^s}(\mathbf{x}^{(i)}) + \mathbf{x}^{(i)}.
\end{equation}

The volume density correction field, noted as $\mathcal{D}^{\sigma}$, is also conditioned on spatial location in individual object space $\mathbf{x}\in\mathcal{O}_i$ and the shape latent $\mathbf{z}_i^s$.

After warping the coordinates and applying volume density correction, we can evaluate the volume density values on spatial location of individual objects' space as follows:
\begin{equation}
\begin{aligned}
    \sigma_i(\mathbf{x}) 
    &= \sigma_{\tau} \left( \mathbf{x}^{(\tau)} \right) + \Delta\sigma \\
    &= \sigma_{\tau}\left( \mathcal{D}^{w}|_{\mathbf{z}_i^s}(\mathbf{x}) + \mathbf{x} \right) + \mathcal{D}^{\sigma}|_{\mathbf{z}_i^s}(\mathbf{x}),
\end{aligned}
\end{equation}
where $\sigma_{\tau}(\cdot)$ denotes the volume density branch of the NeRF template $\mathcal{T}$.

\begin{figure*}[ht]
	\centering
	\includegraphics[width = 1\linewidth]{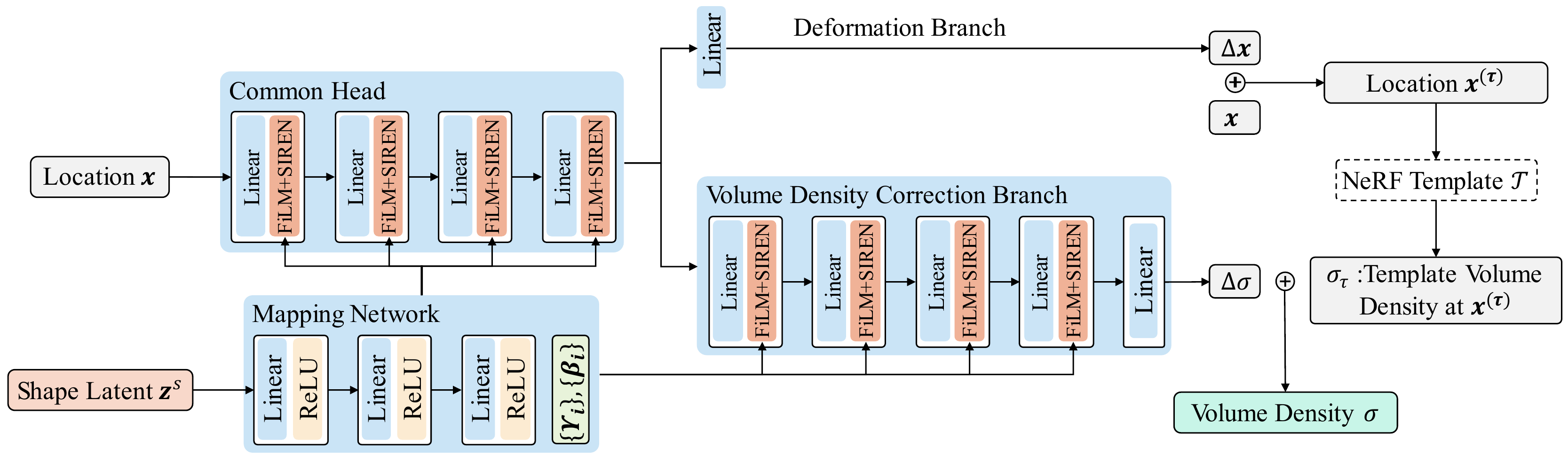}
	\caption{The detailed structure of deformation field and volume density correction field.}
	\label{fig:shape-var}
\end{figure*}

In practice, $\mathcal{D}^{w}$ and $\mathcal{D}^{\sigma}$ share a common head, built with sinusoidal activation and conditioned on the shape latent $\mathbf{z}^{s}$ using the FiLM-SIREN layer as defined in the previous section Eq.~(\ref{equ:film-siren}). 
The structure of the deformation field $\mathcal{D}^{w}$ and the volume density field $\mathcal{D}^{\sigma}$ are shown in Figure~\ref{fig:shape-var}.

\subsection{Learning Appearance, Shape and Correspondences Simultaneously}
To optimize the model parameters, we adopt the following loss and regularization functions~\cite{deng2020DIF, zheng2020DIT, tretschk2020non}.

{\bf Reconstruction Loss.}
We adopt  the standard mean-squared error (MSE) between the ground-truth images ${I}$ and rendered images $\hat{I}$:
\begin{equation}
    \mathcal{L}_{\text{rec}} = \|I-\hat{I}\|_{2}^{2}.
\end{equation}
Learning shapes, appearance, and shape correspondences from images simultaneously is highly under-constrained. 
Thus we apply four regularization terms to stabilize the training process and improve the results.

{\bf Auto-decoder Regularization.}
We use the $L_2$-norm to constrain the learned shape and appearance latent codes into a limited distribution space:
\begin{equation}\label{equ:ad}
    \mathcal{L}_{\text{reg}} = \frac{1}{N} \sum_{i}^N \left( \lVert \mathbf{z}_i^s \rVert_2^2 + \lVert \mathbf{z}_i^a \rVert_2^2 \right).
\end{equation}
Alternatively, the $L_2$-norm regularization can be replaced by minimizing the Kullback–Leibler divergence between the latent code posterior distribution and Gaussian distribution as in VAE~\cite{kingma2013auto} / VAD~\cite{zadeh2019vad} training.

{\bf Correction Regularization.}
It is commonly known that objects from the same category often have relatively few topology variances. To encourage the model to represent shape variances through geometry deformation rather than changing topologies, we penalize the L1-norm of the volume density correction:
\begin{equation}
    \mathcal{L}_{\text{cor}} = \| \mathcal{D}^{\sigma} \|_1.
\end{equation}

{\bf Surface Normal Consistency Regularization.}
Intuitively, objects' corresponding parts should face corresponding directions, i.e. the surface normal in the template space of corresponding points from two objects should be consistent. Hence, we define a normal consistency regularization loss:
\begin{equation}
    \mathcal{L}_{\text{nor}} = 1 - \langle \nabla{\sigma}, \nabla{\sigma_{\tau}} \rangle.
\end{equation}
In DIT~\cite{zheng2020DIT}, since the ground truth shape and ground truth surface normal are known, they can directly encourage the template's surface normal to be consistent with ground-truth object instances' surface normal. In our case, we alternatively encourage the template's surface normal $\nabla{\sigma_{\tau}}$ to be consistent with the predicted surface normal $\nabla{\sigma}$.

{\bf Smoothness Regularization.}
To encourage smooth deformation and avoid large shape distortion, we add a simple smoothness loss on the deformation field:
\begin{equation}
    \mathcal{L}_{\text {smo}}=\left\|\mathbf{J} - \mathbf{I}\right\|_{2},
\end{equation}
where $\mathbf{J} \in \mathbb{R}^{3 \times 3}$ is the Jacobian matrix of deformation field and $\mathbf{I} \in \mathbb{R}^{3 \times 3}$ is an identity matrix.

In summary, the whole training process can be formulated as the following:
\begin{equation}
    \arg \min \mathcal{L}_{\text{rec}} + w_{1} \mathcal{L}_{\text{reg}}  + w_{2} \mathcal{L}_{\text{cor}} + w_{3} \mathcal{L}_{\text{nor}} + w_4 \mathcal{L}_{\text{smo}},
\end{equation}
where $w_{1:4}$ are weights for different loss terms.

\section{Experiments and Analysis}

\begin{figure*}[htbp]
\centering
\includegraphics[width=1\textwidth]{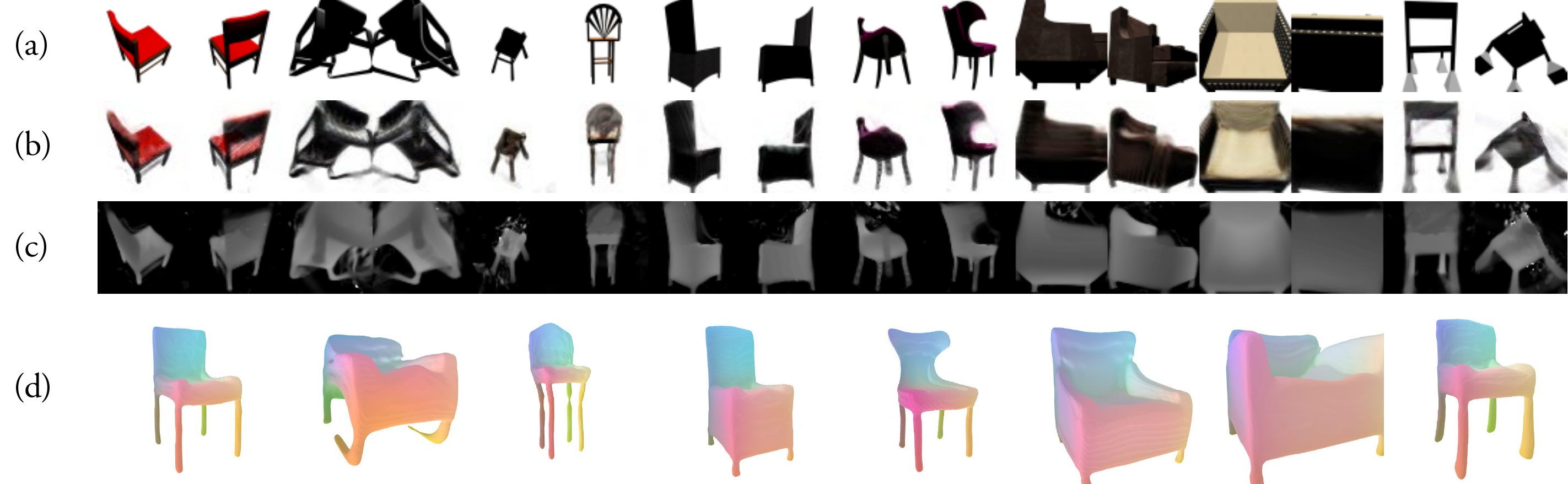}
\caption
    {\textbf{Demonstration of representations learned by Template-NeRF on Shapenet chairs~\cite{chang2015shapenet}}. We select examples from two different views of eight chairs, with (a) ground-truth image, (b) predicted image, (c) predicted depth, and (d) learned shapes and correspondences. Consistent colors across shapes represent the correspondences.}
\label{fig:result}
\end{figure*}

In this section, we first introduce our experiment settings and then evaluate the performance of Template-NeRF qualitatively and quantitatively.

\subsection{Implementation Details}
\label{implementation_details}
We learn Template-NeRF on synthetic renderings provided by~\cite{kato2018neural, sitzmann2019srns} of ShapeNet-v2~\cite{chang2015shapenet}. This dataset consists of a collection of 2D images with camera parameters.
We train our models on a single NVIDIA RTX 3090 GPU which costs 16 hours with a batch size of 5. We use the Adam optimizer with $\beta_1=0, \beta_2=0.9$, We initialize learning rate to $1e{-4}$ and anneal half every 100K steps.
We set other parameters to $w_1=5e{-5}$, $w_2=1e{-3}$, $w_3=5e{-1}$, $w_4=1e{-3}$.

\subsection{Shape Reconstruction}
We first evaluate the power of Template-NeRF by the performance of 3D reconstruction. 
We achieve competitive results with DIF~\cite{deng2020DIF} and DIT~\cite{zheng2020DIT}, with only posed image supervision, as demonstrated in Figure~\ref{fig:result}. 
Note that in Figure~\ref{fig:result}(a) the ground-truth images with poses are the only input, no 3D information is needed during the whole training and testing process. The shapes are extracted using Marching Cubes~\cite{lorensen1987marching} on the volume density predicted by the model in Figure~\ref{fig:result}(d).

\subsection{Shape Correspondences}
Then we evaluate the performance on dense correspondences of our method, as shown in Figure~\ref{fig:teaser}. 
We depict the same correspondence by consistent colors.
To better demonstrate the learned correspondences, we manually select several points on one of the objects and compute their new location on the remaining objects.
The results show that our method can reason dense and semantic correspondences across different objects within the same category. 
Our representation model is also powerful enough to deal with large deformations.
In Figure~\ref{fig:Corr}, we extract a template shape of the chair and establish the coordinate mapping between multiple objects within the same category.

\section{Conclusion and Future Work}
We have presented Template-NeRF, a novel category-specific learning approach to reason dense shape correspondences from pure posed images explicitly. We have introduced the ideas of deep implicit templates to the neural rendering framework NeRF and go after careful designs and modifications, leveraging recent advances in periodic functions and FiLM conditioning. Various experiments demonstrate the great potential of learning dense correspondences directly from images of a wide range of object classes.

{\small
\bibliographystyle{ieee_fullname}
\bibliography{ref}
}
\end{document}